\pdfoutput=1
\documentclass[11pt]{article}

\usepackage{EMNLP2022}

\usepackage{times}
\usepackage{latexsym}
\usepackage{multirow}
\usepackage{graphicx}
\usepackage{wrapfig,lipsum,booktabs}

\usepackage[T1]{fontenc}
\usepackage[utf8]{inputenc}

\usepackage{microtype}

\usepackage{inconsolata}

\usepackage{url}
\usepackage{booktabs}       % professional-quality tables
\usepackage{amsfonts}       % blackboard math symbols
\usepackage{nicefrac}       % compact symbols for 1/2, etc.
\usepackage{microtype}      % microtypography
\usepackage{xcolor}         % colors
\usepackage{algorithm}
\usepackage{listings}

\usepackage{comment}
\usepackage[noend]{algpseudocode}
\usepackage{amsmath}
\usepackage{amsthm}
\usepackage{amsfonts}
\usepackage{dsfont}
\usepackage{multirow}
\usepackage{subfig}
\usepackage{xspace}
\usepackage{wrapfig}
\usepackage{xcolor}
\usepackage{cleveref}
\usepackage{enumitem}
\usepackage{amssymb}
\usepackage{pifont}
\usepackage{caption}
\usepackage{xcolor,pifont}

\setlist[itemize]{noitemsep, nolistsep}

\newcommand{\sjn}[2]{}{}
\newcommand{\name}{PENTATRON}

\newtheorem{definition}{Definition}

\DeclareMathOperator*{\argmax}{argmax} % thin space, limits underneath in displays

\title{\name: PErsonalized coNText-Aware Transformer for Retrieval-based cOnversational uNderstanding}

\author{Niranjan Uma Naresh\thanks{$^*$Equal contribution.}\qquad Ziyan Jiang$^*$ \qquad Ankit$^*$ \\
{\bf Sungjin Lee} \qquad {\bf Jie Hao} \qquad {\bf Xing Fan} \qquad {\bf Chenlei Guo} \\
Amazon \\
\small \texttt{\{niumanar, ziyjiang, ankitvys, sungjinl, jieha, fanxing, guochenl\}@amazon.com}
}

\begin{document}
\maketitle

\begin{abstract}
Conversational understanding is an integral part of modern intelligent devices. In a large fraction of the global traffic from people using smart digital assistants, frictions in dialogues may be attributed to incorrect understanding of the entities in a user's query due to factors including ambiguous mentions, mispronunciation, background noise and faulty on-device signal processing\sjn{ambiguous mentions?}{added}. Such errors are compounded by two common deficiencies\sjn{I would drop ``of responses''}{removed} from intelligent devices namely, (1) the device not being \sjn{how about ``the device not being''?}{changed} tailored to individual users, and (2) the device responses being unaware of the context in the conversation session \sjn{Can we clarify what kind of context?}{sure, added ``in-session''}. Viewing this problem via the lens of retrieval-based search engines, we build and evaluate a scalable entity correction system, \name. The system leverages a parametric transformer-based language model to learn patterns from in-session user-device interactions coupled with a non-parametric personalized entity index to compute the correct query, which aids downstream components in reasoning about the best response. In addition to establishing baselines and demonstrating the value of personalized and context-aware systems, we use multitasking to learn the domain of the correct entity. We also investigate the utility of language model prompts. Through extensive experiments, we show a significant upward movement of the key metric (Exact Match) by up to 500.97\% (relative to the baseline). %847bps. %. {\todona add numbers here}
\end{abstract}

%a customer's in-session conversational behavior and responses from the agent from the multi-turn dialogue session to replace incorrect entities in the current query

% Query rewriting (QR) is important in reducing friction in conversational agents like Alexa, Siri and Google Assistant.
% Thus, QR task can be treated as an entity rewriting problem.
% in this paper

%Viewing this problem via the lens of retrieval-based search systems, we propose a personalized contextual entity rewriting model which uses previous queries and responses from the agent from the multi-turn dialogue session to replace incorrect entities in the current query. The correct entities are chosen either from the personalized index of each customer for enhanced personalization and customer satisfaction. We demonstrate how to use context and train our model in single task context to rewrite entity setting as well as multi-task setting by additionally utilizing the Natural Language Understanding (NLU) hypothesis. Our results demonstrate that contextual information within the same session improves the accuracy of query rewriting while also facilitating domain correction.

%In a large fraction of the global traffic from customers using smart digital assistants, frictions in dialogues occur due to incorrect understanding of the entities in the user's query. 

\section{Introduction}
Intelligent devices are ubiquitous in the modern computing. The scientific modules that drive these devices involve conversational understanding, ambient computing, natural language reasoning and self-learning~\cite{thoppilan2022lamda,sarikaya2022intelligent,pinhanez2021using,liu2021pre}. A user's interaction with a device, however, is susceptible to errors arising from a myriad of sources including wrong pronunciation, inaccuracies in the subject mentions in a sentence, environmental noise, hardware and software error~\cite{kim2020review}. Correct interpretations of user queries, especially entities, is central to delivering the best user experience. Two important factors that contribute strongly to high-precision entity \sjn{not sure if ``entity'' is a more accurate term for our use case.}{Good point, edited} recognition are (1) personalization, ie, learning users' unique patterns, and (2) contextualization, ie, deriving cues from the information in a user-device interaction session. In this paper, we design and evaluate an entity correction system, \name, with both personalization and contextualization baked into its architecture.
\begin{figure}[h]
    \centering
    \includegraphics[width=7.5cm]{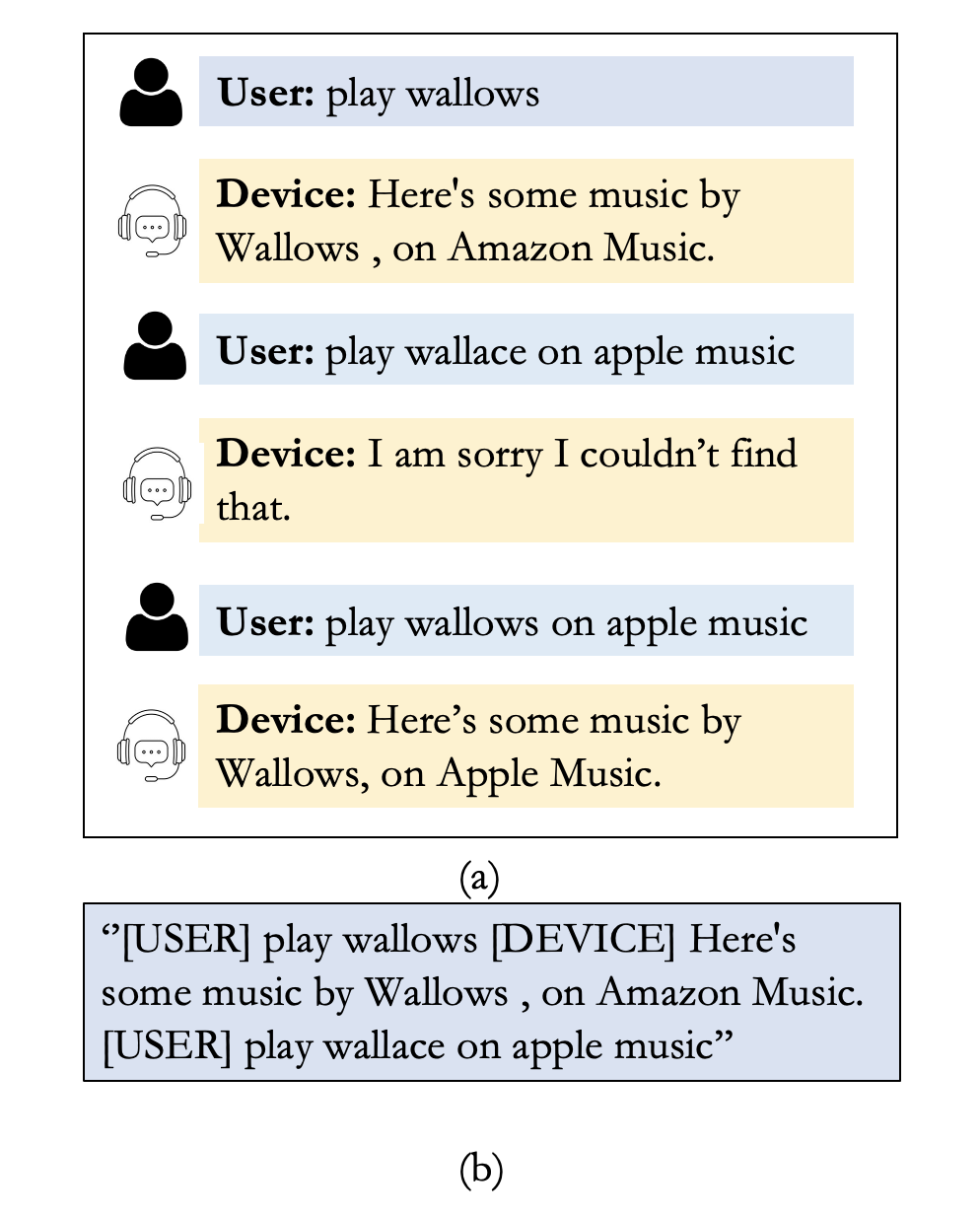}
    \caption{(Above) One multi-turn dialogue session with defective source query which contains one erroneous entity `wallace' and its successful rephrase with correct entity `wallows'. (Below) Concatenation of queries and responses using special tokens to form a single sequence as encoder input.}
    \label{fig:personal_contextual}
\end{figure}

\subsection{Motivation}
In Figure~\ref{fig:personal_contextual}, we illustrate a real-world case as to why personalization and contextualization are very important, especially due to the specificity in highly entity-centric domains such as music. In this case, masking the very last device response, we observe that there is valuable information scattered across the user's requests in the session yet, the device delivers sub-par experience by responding defectively multiple times before finally getting the user's intent right.

\subsection{Notation and Preliminaries}
\begin{definition}
\label{def:nlu_hypothesis}
Let integer $\gamma$ satisfy $1 \leq \gamma < \infty$. A natural language (NL) hypothesis is a mapping, $h: Q \rightarrow D \times I \times [E]^\gamma$, where $Q$ refers to the query space, $D$ refers to the domain space, $I$ refers to the intent space and $E$ refers to the entity space. The entity space, $E:= E_T \times E_V$, may further be decomposed into the entity type space $E_T$ and the entity value space $E_V$. All spaces are defined over Unicode strings.
\end{definition}
As an example, given a query string $q=$``\textit{play the real slim shady}'', the corresponding NL hypothesis is $h(q)=$\textit{(Music, PlayMusicIntent, [(SongName, the real slim shady)])} where the domain is \textit{Music}, the intent is \textit{PlayMusicIntent}, and the entity value is \textit{the real slim shady} with \textit{SongName} entity type.
%As an example, given a query string $q=$``\textit{play the real slim shady by eminem}'', the corresponding NL hypothesis is $h(q)=$\textit{(Music, PlayMusicIntent, [(SongName, example song), (ArtistName, example artist)])} where the domain is \textit{Music} and the intent is \textit{PlayMusicIntent}, and there are two entities namely, the \textit{the real slim shady} entity value with \textit{SongName} entity type and the \textit{eminem} entity value with the \textit{ArtistName} entity type.
\begin{definition}
\label{def:pentatron}
Building on Definition~\ref{def:nlu_hypothesis}, our system, \name, may be formalized as $\Phi:(C,Q) \rightarrow E_V$ where $C$ is the user space (anonymized using a hash function, for privacy, in practice).
\end{definition}
In a nutshell, given an input query $q$ (with or without dialogue context), our system essentially solves the optimization problem,
\begin{align}
\label{eqn:main}
\min_\theta \mathbb{E}_{(c, q, e) \sim \mathcal{D}} \left[ \ell \left( \Phi_\theta(c,q), e \right) \right]
\end{align}
where $\mathcal{D}$ is supported on $C \times Q \times E_V$.
% In contrast, non-personalized systems solve
% \begin{align}
% \min_\theta \mathbb{E}_{(c, q_{in}, q_{out}) \sim \mathcal{D}} \left[ \ell \left( \Phi_\theta(c,q_{in}), q_{out} \right) \right]
% \end{align}

\sjn{Up to this point, the most important two terms ``context'' and ``personal entities" are not clearly defined, leaving some statements confusing. Can we give a more clearer meaning somewhere in this section?}{Agree, just took over the rest of this section and refactored it.}

\subsection{Our Contributions and Preview of Results}

\begin{figure}[h]
    \centering
    \includegraphics[width=\columnwidth]{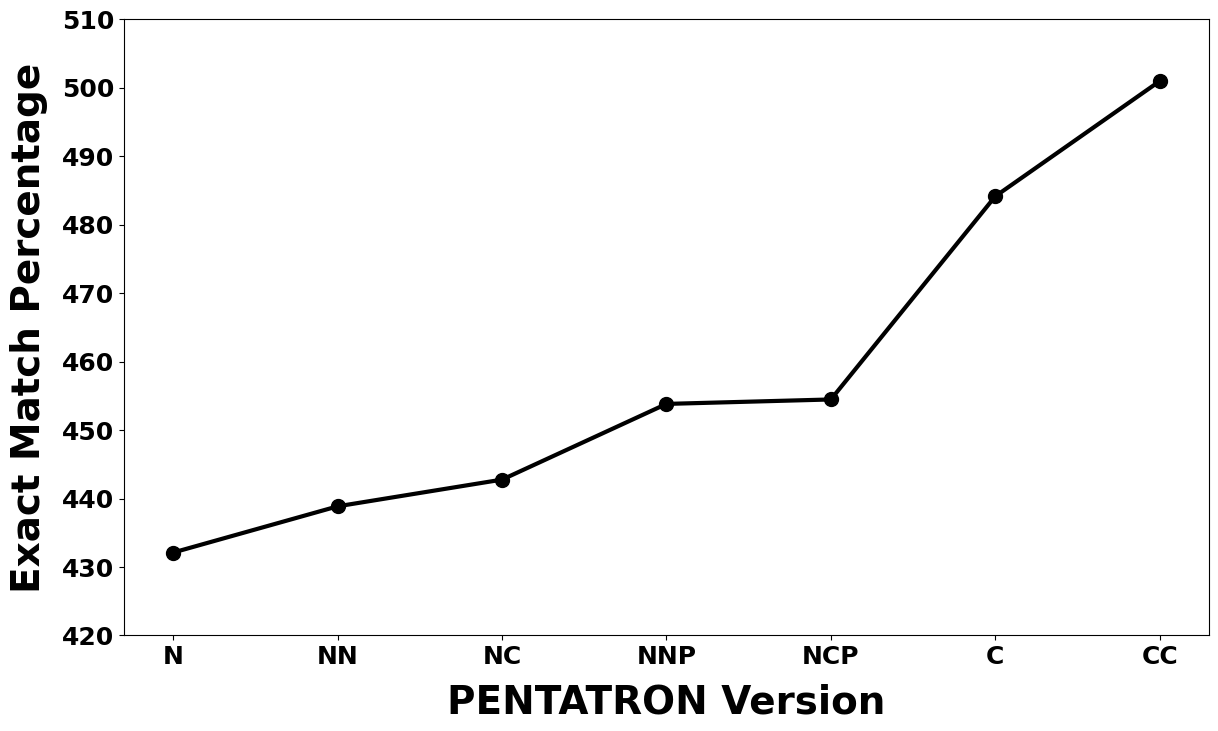}
    \caption{Preview of the system performance which shows consistent significant improvement in going from a purely personalized system (N) to a fully contextual personalized system (CC). Further details are available in Table~\ref{tab:ablations}.}
    \label{fig:accuracy_pentatron}
\end{figure}

%rate}[nolistsep,noitemsep]
%\item \textit{Data analysis: }We use historical multi-turn dialogue logs from English-speaking customers in the United States to conduct analysis.
On the system design front, we build a retrieval-based pipeline\sjn{Does this mean there was no prior work using retrieval-based pipeline? If not, can we elaborate our novelty here?}{Good point, this is not a theory paper, so better not to itemize}. Our model backbone is inspired by attention-based~\cite{vaswani2017attention} transformer encoders~\cite{devlin2018bert}. We achieve personalization via a non-parametric index which is essentially a key-value pair look-up table with the keys representing users and values representing the entity lists derived from historical data aggregation. With respect to experimental results, we conduct extensive studies on seven different versions of \name{}, involving ablations with prompts, multi-tasking and non-contextual training data, and show consistent improvements in Exact Match (EM) of up to 500.97\% (relative to the baseline) as captured by the preview of results in Figure~\ref{fig:accuracy_pentatron}.

%$847$ basis points as captured by the preview of results in Figure~\ref{fig:accuracy_pentatron}. % {\todona add numbers here}

%\end{enumerate}

% we prove that personalization achieves better convergence

%Our results show that using the contextual information improves the accuracy on the non-contextual baseline by 6\% (65.45\% to 71.85\%). The multi-task training further improves the accuracy by 1.52\% (71.85\% to 73.37\%). We also train the multi-task models using prompts for rewrite and domain tasks and get a further improvement in accuracy by <>.

\section{Background and Related Work}
% \begin{table*}[h]
% \caption{Highlights our system with respect to closely-related works in the literature.}
% \label{tab:literature}{}
% \begin{tabular}{lllll}\\\toprule
% \textbf{System} & \textbf{Entity Correction} & \textbf{Personalization} & \textbf{Contextualization} & \textbf{Scalability} \\ \toprule
% BLINK~\cite{wu2019scalable} & \cmark & \xmark & \xmark & \xmark \\  %\midrule
% ELQ~\cite{li2020efficient} & \cmark & \xmark & \xmark & \cmark \\  %\midrule
% CRD~\cite{wang2021contextual} & \xmark & \xmark & \cmark & \xmark \\  \midrule
% This Work & \cmark & \cmark & \cmark & \cmark \\  \bottomrule
% \end{tabular}
% \centering
% \end{table*}

%\subsection{Related Work}

\subsection{Query Rewriting}
Query Rewriting (QR) in dialogue systems aims to reduce frictions by reformulating the automatic speech recognition component's interpretation of users' queries. Initial efforts \citep{dehghani2017learning,su2019improving} treat QR as a text generation problem.

Some recent studies \citep{Chen2020,Yuan2021,Fan2021,cho2021personalized} are based on neural retrieval systems. In the retrieval-based systems, the rewrite candidate pool is aggregated from users' habitual or historical queries so that the rewrite quality can be tightly controlled. Compared to generation-based systems, retrieval-based systems may sacrifice flexibility and diversity of the rewrites, but in the meanwhile provide more stability which is more important in a runtime production setup. 

\textbf{Personalization} and \textbf{Contextualization} are two popular directions for QR systems. A personalized system such as \citealp{cho2021personalized} tends to incorporate diverse affinities and personal preferences to provide individually tailored user experience in a single unified system. Contextualization attempts to utilize multi-turn queries rather than only leveraging single-turn information. Some previous studies \citep{WangZ2021} have shown the benefits by leveraging the dialogue context and user-device interaction signals.

Entities have been shown to be a strong indicator of text semantics. Since queries in our dialogue system are typically short sentences, entities are even more important in this scenario. Most existing QR approaches mentioned above rephrase query utterances entirely. Although some existing works focus on specific categories like coreference resolution or entity omission \citep{su2019improving,tseng2021cread}, none of them has a particular emphasis on the correction of erroneous entities.

\subsection{Entity Linking}

Another related thread towards our task is entity linking. Entity linking task aims to link mentioned entities with their corresponding entities in a knowledge base. In a retrieval-based QR system which focuses on entity correction, we could adopt similar methods in entity linking area. BLINK\citep{wu2019scalable} designs a two-stage retrive-rerank framework based on pre-trained deep transformers. The following work ELQ \citep{li2020efficient} uses a biencoder to jointly perform mention detection and linking in one pass and also shows good improvement in latency metrics which is quite important in a production settings. Our task is more challenging than entity linking because the input utterance is noisy with incorrect entities and the lack of textual descriptions of each entity. 

%Contextual Rephrase Detection (CRD), as designed in \cite{wang2021contextual}, implements an offline system that goes beyond a pairwise learning system. They mine conversational texts from historical multi-turn dialogues and propose the extraction of relevant rephrases using a classification approach.

%\subsubsection{Prompt tuning}
%Attention models have been applied in context-aware ranking systems have shown good empirical performance~\cite{pobrotyn2020context}. Prompt-based approaches have shown promising empirical results in various natural language applications~\cite{liu2021pretrain}. Relevance models have a rich history in search systems~\cite{he2016learning}.
%

%latency and production and multitasking span is serparte

\section{Problem Setup and Solution Design}

\begin{figure}[h]
    \centering
    \includegraphics[width=7.5cm]{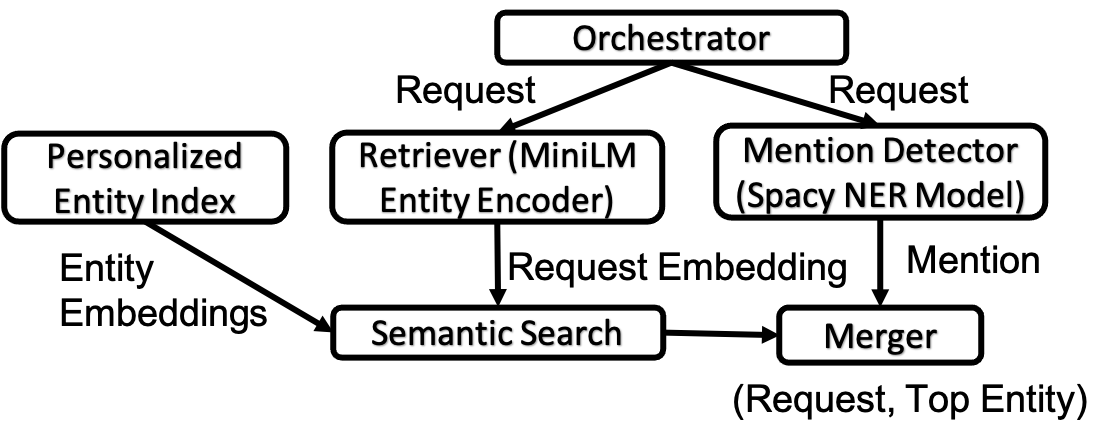}
    \caption{For a given user, the input request string from the \name{} orchestrator is processed by a transformer model and also by a named entity recognition model, both trained on historical user requests, to encode the request and extract mentions, respectively. A semantic search is applied on the request embeddings and the precomputed entity embeddings of the user to find the best match following which, post-processing is applied to feed the result into downstream components.}
    \label{fig:pentatron_architecture}
\end{figure}

The overall architecture of the \name{} system is described in Figure~\ref{fig:pentatron_architecture}.

\subsection{Entity Correction in Query Rewriting}
% We frame the query rewrite problem as an information retrieval task.
We consider a dataset of $M$ multi-turn dialogue sessions: $\{S_t\}_{t=1}^T$. $S$ is a set of $T$ turns in chronological order: $S = \{ (q_t, r_t) \}_{t=1}^T$. Here $t$ is the index of turn and each turn consists of a pair $(q_t, r_t)$, where $q_t$ denotes the user's query utterance and $r_t$ denotes the device's response utterance. The sessions are selected so that the source query $q_{T-1}$ contains one erroneous entity and $q_{T}$, which has the correct form of entity $e$, is the rephrase of the previous turn. More details about the data selection is described in Section 4.1. Our prediction goal is formulated as:
\begin{align}
    \widehat{e} & = \argmax_e \mathbf{P}(e \mid \{S_t\}_{t=1}^{T-2}, q_{T-1}) \\
    \widehat{q}_T & = g(q_{T-1}, \widehat{e})  
\end{align}

We flatten the previous dialogue turns $\{S_t\}_1^{T-2}$ and the source query $q_{T-1}$ into a single sequence to feed into the encoder, as shown in Figure~\ref{fig:personal_contextual}.
Since the only difference between $q_{T-1}$ and $q_{T}$ is whether we have the correct form of $e$, the final rewrite is generated based on source query $q_{T-1}$ and entity prediction $\hat{e}$ through a simple replacement function $g$.

%The entities that are scored are selected only from the personalized index of the customer. The entity rewrite process is triggered based on the score of the top-2 entities.

% We use a similar way as entity layer of personalized index mentioned in \cite{cho2021personalized}. 

% To construct a personalized entity index, 

\subsection{Personalized Entity Index}
We build an personalized entity index for each user to leverage individual interaction history by aggregating users' frequent entities in past 30 days\footnote{All user information is in a de-identified format.}. The entities include song names or artists that users frequently listened to, nicknames of users' intelligent devices and so on.

This index serves as the retrieval candidate pool during inference time. The candidate embeddings are cached. We implement a two-stage in-memory index that has a map of users to their specific entities along with the embeddings corresponding to the union of entities across all users. This is done for memory efficiency reasons so that we avoid the overhead caused by the redundancy of duplicate entities across different users.

\subsection{Modeling}
%We use a bi-encoder architecture to model compute the query and the entity embeddings. We follow the method proposed in \citealp{wang2021contextual} to utilize dialogue session data, ie, we flatten the dialogue session into one sequence.

%\subsection{single task}
%2
%\subsection{multitask without prompt}
%4
%\subsection{multitask with prompt}
%4

We use a bi-encoder architecture based on MiniLM~\cite{wang2020minilm} for jointly encoding the queries and the entities~\cite{humeau2019poly}. The weights are shared for memory footprint savings and serving cost reduction. Note, we also try asymmetric query and candidate entity encoders; however, we observe only a marginal performance improvement of less than $1\%$. %The encoder model of the \name{} system is trained with the primary task of maximizing the similarity score of the input (user query with or without context) and the target correct entities, operating within the randomly selected batch.% 
We use a batch size of $128$ and train it on p3.2x-large GPU instances %~\footnote{Our GPU instance is equipped with 16GB on-device memory}
acquired on AWS cloud. AdamW~\cite{loshchilov2017decoupled} is our optimizer of choice.

For detecting mentions in the input query, we use a Spacy named entity recognition model trained on historical user queries containing entity strings from different domains.

% Emb_context = Transformer1(context)
% Emb_entity_i = Transformer2(entity_i)

% Score(context, entity_i) = Emb_context . Emb_entity_i / (length of Emb_context)*( length of Emb_entity_i)

%A semantic search then finds the similarity scores based on cosine similarity and retrieves the top-k <value of k> candidates. In the next stage, we utilize a trigger threshold for triggering the rewrite. If the cosine similarity of top-1 entity is greater than threshold 1, and that of the top-2 is lower than threshold 2, the model rewrites the entity in the query with the top-1 entity. This rewrite entity is then matched with the gold entity to compute the evaluation metrics.

%\subsection{Training Schemes}

% \section{Solution Design} % Methodology
%In this section, we first describe our transformers-based retrieval approach followed by an explanation of the multiple training paradigms that we utilize.

\subsection{Optimization Objectives}
A combination of both hard negatives~\cite{gillick2019learning} and in-batch random negatives improve the performance of large-scale natural language reasoning systems.
We use the multiple negatives ranking loss~\cite{henderson2017efficient} for the primary task. We take a metric learning approach~\cite{hadsell2006dimensionality} to the auxiliary task, ie, we use the contrastive loss here. %In this section, we use $f$ denote our encoder model. Specifically, we consider the embedding of it's [CLS] token in the final layer.

\paragraph{Inference: }The semantic search function which is used in primary retrieval task computes $s_i = \cos(f(q), f(e_i))$ for $i \in [k]$ where $e_i \in E_V$ are the top-$k$ entities retrieved from the personalized index (sorted by the relevance score in descending order) and $q$ refers to the query (with or without context). We configure our system to be activated on the threshold conditions, $s_1 > \tau_1$ and $s_2 < \tau_2$, to make sure the top-$2$ entities are sufficiently far apart to avoid any ambiguous predictions.% For the auxiliary retrieval task, we compute scores $v_j= \cos(f(q), f(d_j))$, where $j$ indexes over the domain strings, and pick the domain $d_1$ with the top score.

\paragraph{Training: } The encoder model of the \name{} system is trained with the primary task of entity prediction, which we maximize the similarity score between the user query (with or without context) and the target correct entity.
Consider a batch of $N$ samples. The loss of the primary task is given by:
\begin{align}
    \mathcal{L}_E = - \frac{1}{N} \sum_{i=1}^N \log \frac{\exp(s_i)}{\sum_{j=1}^N \exp(s_j)}
\end{align}
%h(q)
In the above formula, we only take in-batch random negatives into consideration. We will also discuss the utilization of hard negatives later in this paper.

%In live traffic, one dialogue session could incorporate several actions from the user. For example, the user wants to turn on the bedroom light and play some music in one dialogue session. In this situation, unrelated turns from the context could introduce additional noise and influence the performance of the system. To mitigate this, we adopt an auxiliary task during training, which we called domain-aware query matching. In this task, we want to push source queries targeting to the same domain close to each other.

We adopt an auxiliary task during training to have an implicit clustering effect of the query embeddings based on target domain. For this task, we want to push source queries targeting to the same domain close to each other and source queries targeting to the different domain away from each other.

For $N$ randomly selected pairs of queries (indexed by $i$ and $j$) from a batch, the loss of the auxiliary task is the contrastive loss given by:
% \begin{align}
%     \mathcal{L}_D = & \frac{1}{N} \sum_{(i,j)} \mathbf{1} \{h^D(q_i) = h^D(q_j)\}.(v_i - v_j)^2 \nonumber \\
%     & + \frac{1}{N} \sum_{(i,j)} \mathbf{1} \{h^D(q_i) \neq h^D(q_j)\} . \nonumber \\
%     & \max(0, \lambda - (v_i - v_j)^2)
% \end{align}
\begin{align}
    \mathcal{L}_D = & \frac{1}{N} \sum_{(i,j)} \mathbf{1} \{h^D(q_i) = h^D(q_j)\} . \nonumber \\
    & \| f(q_i) - f(q_j) \|^2 \nonumber \\
    & + \frac{1}{N} \sum_{(i,j)} \mathbf{1} \{h^D(q_i) \neq h^D(q_j)\} . \nonumber \\
    & \max(0, \lambda - \| f(q_i) - f(q_j) \|^2)
\end{align}
The margin parameter $\lambda$ is set as $0.75$. Here, $h^D$ denotes the domain extracted from the NL hypothesis of the target (final) dialogue turn.

\paragraph{Multi-task Formulation: }The final loss is computed as $\mathcal \mu \mathcal{L}_E + (1-\mu) \mathcal{L}_D$ where $\mu \in (0,1]$. Specifically, we build different versions of \name{} by setting $\mu = 1$ and $\mu = 0.5$. %Using this, we observe an implicit domain-wise clustering effect of requests.

%\subsubsection{Single task baselines}
We train two single-task models which are used as the non-contextual and contextual baselines respectively. The non-contextual baseline model uses the source query as input and the rewrite entity as output. The contextual baseline model uses the full context (truncated to maximum allowable length of $256$) as input and the rewrite entity as output.

% \begin{table*}[h]
% \centering
% \begin{tabular}{lllll}
% \toprule
% \bf{Model} & \bf{Task} & \bf{Trigger} & \bf{Accuracy} \\
% \toprule
% \name$_{rr}$ & request to rewrite & & \\  % model 1
% \name$_{cr}$ & context to rewrite & & \\ % model 2
% \name$_{rr}$ & request to rewrite & & \\  % model 3
% \name$_{cr}$ & context to rewrite & & \\ % model 4
% \name$_{rr}$ & request to rewrite & & \\  % model 5
% \name$_{cr}$ & context to rewrite & & \\ % model 6
% \name$_{rr}$ & request to rewrite & & \\  % model 7
% \name$_{cr}$ & context to rewrite & & \\ % model 8
% \name$_{rr}$ & request to rewrite & & \\  % model 9
% \name$_{cr}$ & context to rewrite & & \\ % model 10
% \bottomrule
% \end{tabular}
% %\caption{Single task non-contextual (Model 1) and contextual (model 2) baselines}
% \caption{List of all model settings and their performance numbers}
% \label{tab:ablations}
% \end{table*}

\begin{table*}[h]
\centering
\begin{tabular}{lllll}
\toprule
\bf{Model} & \bf{Primary Task} & \bf{Auxiliary Task} & \bf{Exact Match (Relative)} \\
\toprule
DPR-EC & Non-contextual & None & 0.0 [Baseline] \\ % 12.30\%  \\ % global level retrieval
\midrule
\name-N & Non-contextual & None & +432.11\% \\ % 65.45\% \\  % model 1
\name-NN & Non-contextual & Non-contextual & +438.86\% \\ % \\ 66.28\% \\  % model 3
%\name$_{cr}^{rd}$ & context to rewrite & request to domain & 70.38\% \\ % model 4
\name-NC & Non-contextual & Contextual & +442.76\% \\ % 66.76\% \\  % model 5
%\name$_{cr}^{cd}$ & context to rewrite & context to domain & 73.92\% \\ % model 6
\name-NNP & Non-contextual & Non-contextual with prompt & +453.82\% \\ % 68.12\% \\  % model 7
%\name$_{cr}^{prd}$ & context to rewrite & prompt + request to domain & 72.21\% \\ % model 8
\name-NCP & Non-contextual & Contextual with prompt & +454.47\% \\ % 68.20\% \\  % model 9
%\name$_{cr}^{pcd}$ & context to rewrite & prompt + context to domain & 71.24\% \\ % model 10
\name-C & Contextual & None & +484.14\% \\ % 71.85\% \\ % model 2
\name-CC & Contextual & Contextual & +500.97\% \\ % 73.92\% \\ % model 6
\bottomrule
\end{tabular}
%\caption{Single task non-contextual (Model 1) and contextual (model 2) baselines}
\caption{List of all model settings and their performance numbers (relative, with respect to the baseline, DPR-EC). The primary task is entity prediction using the multiple negative ranking loss with a batch size of $128$ and the auxiliary task uses the online contrastive loss with a margin of $0.75$. We apply the state-of-the-art retrieval model, DPR~\cite{karpukhin2020dense}, to train a dual BERT architecture, DPR-EC, for entity correction as the baseline, i.e., without utilizing personal and contextual information.}
\label{tab:ablations}
\end{table*}

% \begin{table*}[h]
% \centering
% \begin{tabular}{lllll}
% \toprule
% \bf{Model} & \bf{Task1} & \bf{Task2} & \bf{Accuracy} & \bf{Trigger} \\
% \toprule
% \name$_{rr}$ & request to rewrite & --- & 65.45\% & 2.74\% \\  % model 1
% \name$_{cr}$ & context to rewrite & --- & 71.85\% & 3.11\%\\ % model 2
% \name$_{rr}^{rd}$ & request to rewrite & request to domain & 66.28\%  & 4.65\% \\  % model 3
% \name$_{cr}^{rd}$ & context to rewrite & request to domain & 70.38\%  & 3.14\% \\ % model 4
% \name$_{rr}^{cd}$ & request to rewrite & context to domain & 66.76\%  & 2.82\% \\  % model 5
% \name$_{cr}^{cd}$ & context to rewrite & context to domain & 73.92\%  & 3.41\% \\ % model 6
% \name$_{rr}^{prd}$ & request to rewrite & prompt + request to domain & 68.12\%  & 2.45\% \\  % model 7
% \name$_{cr}^{prd}$ & context to rewrite & prompt + request to domain & 72.21\%  & 3.12\% \\ % model 8
% \name$_{rr}^{pcd}$ & request to rewrite & prompt + context to domain & 68.20\%  & 2.29\% \\  % model 9
% \name$_{cr}^{pcd}$ & context to rewrite & prompt + context to domain & 71.24\%  & 2.91\% \\ % model 10
% \bottomrule
% \end{tabular}
% %\caption{Single task non-contextual (Model 1) and contextual (model 2) baselines}
% \caption{List of all model settings and their performance numbers}
% \label{tab:ablations}
% \end{table*}

%\subsubsection{Multitasking}
Furthermore, we train another five versions \name{} with multi-task settings. We also investigate the usage of task markers similar to the approach in ~\citealp{maillard2021multi}. The (hard) prompts are added as special tokens [REWRITE] and [DOMAIN] before the corresponding input during training. \sjn{It's well known that the performance of prompt-tuning varies quite a lot according to the prompt. The special symbol-based prompts may not be the best to leverage the semantics encoded in natural language. Have you tried natural language-based prompts?}{Yes, great point. Haven't had the time to do NL prompting but we should consider it in upcoming iterations} More details are presented in Table~\ref{tab:ablations}\sjn{The combination is not exhaustive. Any reason?}{Some of the results are inconsistent (eg, trigger thresholds not tuned, accuracy and trigger rates are zig-zag, pr-auc has not been computed, etc)}

\paragraph{Hard Negatives Mining: }First, we use bm25 ~\cite{robertson2009probabilistic} to mine hard negatives from the candidate pool, which shows minor improvement. Hence, we adopt a two-pass method to compute hard negatives. In the first pass, we use a model trained with random negatives to perform inference on a disjoint ``second'' training set to obtain entity predictions. In the second pass, we continue training the previous baseline model checkpoint and take into account the wrong predictions as hard negatives.

% bm25

% \name$_{00}$
% \begin{table}[h]
% \centering
% \begin{tabular}{llll} 
% \toprule
% \bf{Model} & \bf{Task 1} & \bf{Task 2} \\
% \toprule
% 3/7 & request to rewrite  & request to domain \\ 
% 4/8 & context to rewrite  & request to domain \\ 
% 5/9 & request to rewrite  & context to domain \\
% 6/10 & context to rewrite  & context to domain \\
% \bottomrule
% \end{tabular}
% \caption{Multitask models with/without prompts}
% \end{table}

%\subsection{Optimization}

\section{Experiments}
\subsection{Training and Test Data}
\label{sec:train_test_data}
% \begin{table}[h]
% \centering
% \begin{tabular}{lll}
% \hline
% \textbf{} & \textbf{Train} & \textbf{Test}\\
% \hline
% \#samples & {2026577} & {200015}\\
% \#users & {1501705} & {181870} \\
% \hline
% \end{tabular}
% \caption{Statistics of training and test data}
% \label{tab:train_test_statistics}
% \end{table}

Our data is derived from the logs of a commercial voice assistant and we process the data with strict privacy standards so that users are not identifiable. We sample multi-turn dialogue sessions between English-speaking users and devices in a time period of one month, in May-June 2022, from all over the United States. A defect detection model similar to \cite{gupta2021robertaiq} and rule-based filters are applied to find dialogue sessions whose last two turns of user query are rephrase pairs. Rule-based filters are using edit-distance and time gap between utterance pairs similar to \cite{cho2021personalized}. Since our work has a particular emphasis on the correction of erroneous entities, we also utilize the NL hypothesis of the rephrase pairs to get such cases.
%<1-2 lines on the basis of which the sessions are selected>.
For simplicity, we consider data with only a single erroneous entity as the target to be predicted. It is straightforward to generalize our system to the multiple entities case. 
%The previous queries generally contain an incorrect rephrase of what the customer is trying to achieve, see image <> for an example. Hence, the input to our model is the following: $[[USER] + [RESPONSE]]_{i=1}^{n-1} + [USER]$.

We sample the test set and keep only sessions wherein a retrieval-based system such as \cite{cho2021personalized}, which rephrases query utterances entirely, couldn't solve. For training, a sample of two million utterances was extracted. Also, some (completely generic) example dialogs extracted from critical data are reported in the paper (Table~\ref{tab:contextual_case_study}).
%Table~\ref{tab:train_test_statistics} contains more information about the training and test sets.

\begin{figure}[h]
    \centering
    \includegraphics[width=\columnwidth]{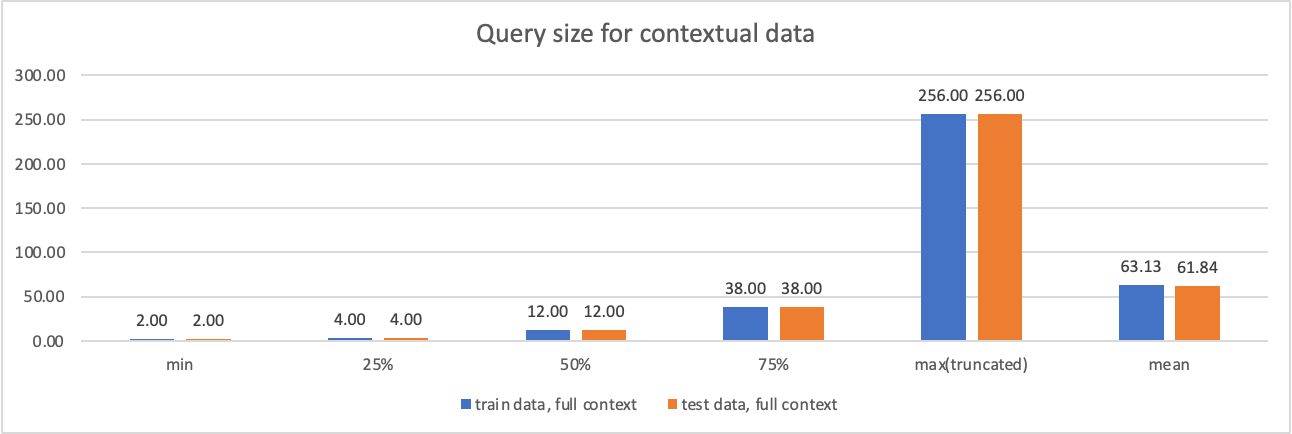}
    \caption{Query length statistics of contextual training and test data.}
    \label{fig:query_length_contextual}
\end{figure}

\begin{figure}[h]
    \centering
    \includegraphics[width=\columnwidth]{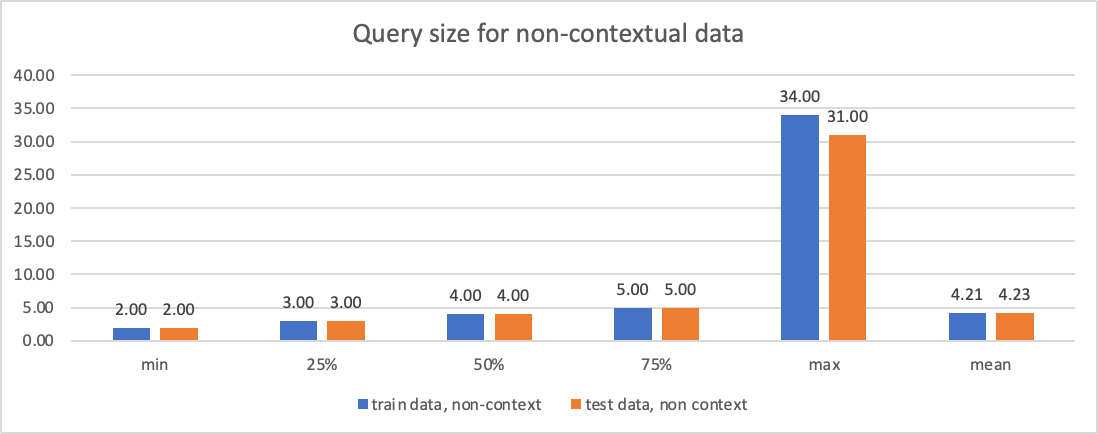}
    \caption{Query length statistics of non-contextual training and test data.}
    \label{fig:query_length_noncontextual}
\end{figure}

Figures \ref{fig:query_length_contextual} and \ref{fig:query_length_noncontextual} summarize the keys data statistics on the training and test sets. This gives us an insight into how transformer models stand to benefit from longer sequences in our application since they are parameterized by and compute second-order statistics.

\subsection{Evaluation Metrics}
We utilize the harshest metric to evaluate our system namely, the Exact Match (EM). This score is 1 if the predicted rewrite exactly matches the labeled rephrase, and is 0 otherwise. We use the same threshold $\tau_1$ and $\tau_2$ for all the proposed \name{} models. The threshold is experimentally set up to keep the balance between opportunities and potential risks in real production.

%i) {\bf{Accuracy}}: Accuracy is the \% of rewrites which are correct. We consider overall accuracy (with inter-domain movements allowed) and intra-domain accuracy.\\
%ii) {\bf{Trigger rate}}: Trigger rate is the \% of samples for which the model triggers a rewrite based on the trigger threshold 1 and trigger threshold 2.\\

%\subsection{Baseline and Experimental Setup}
%To evaluate whether contextual information and multi-tasking add value in identifying correct entities that aid in producing good rewrites, we utilize a non-contextual personalized model as the baseline. We fine-tune the original MiniLM model on the non-contextual training dataset with the last query as the input and rewrite entity as the output. 

%\footnote{https://huggingface.co/sentence-transformers/all-MiniLM-L6-v2}

%\subsection{Ablations}
%We build different versions of systems as shown in Table~\ref{tab:ablations}. Figure~\ref{fig:thresholding} presents an ablation that shows the benefits of hard negative sampling. To further stress test our system, we also swept over random initializations and different thresholds, summarized in Figure~\ref{fig:accuracy_vs_threshold}.

\begin{figure}[h]
    \centering
    \includegraphics[width=\columnwidth]{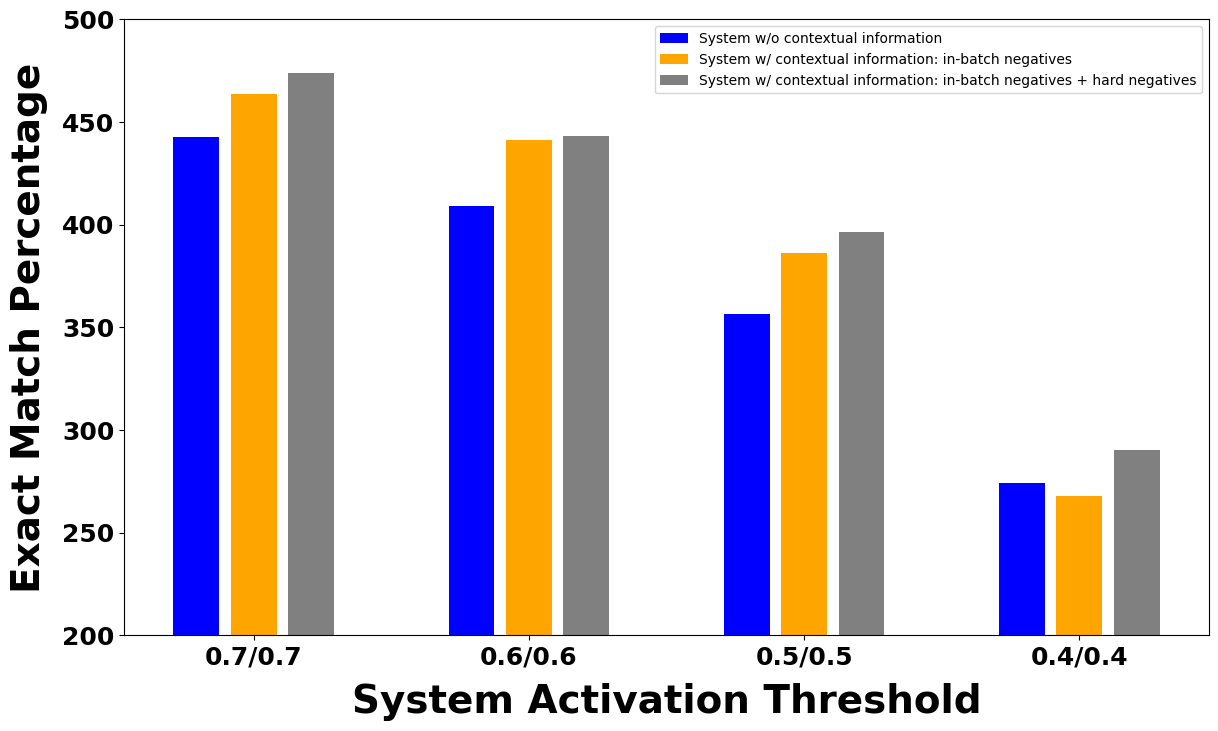}
    \caption{In this figure, we illustrate the importance of contextual information and training with hard negatives in boosting the performance of our system.}
    \label{fig:thresholding}
\end{figure}

\begin{figure}[h]
    \centering
    \includegraphics[width=\columnwidth]{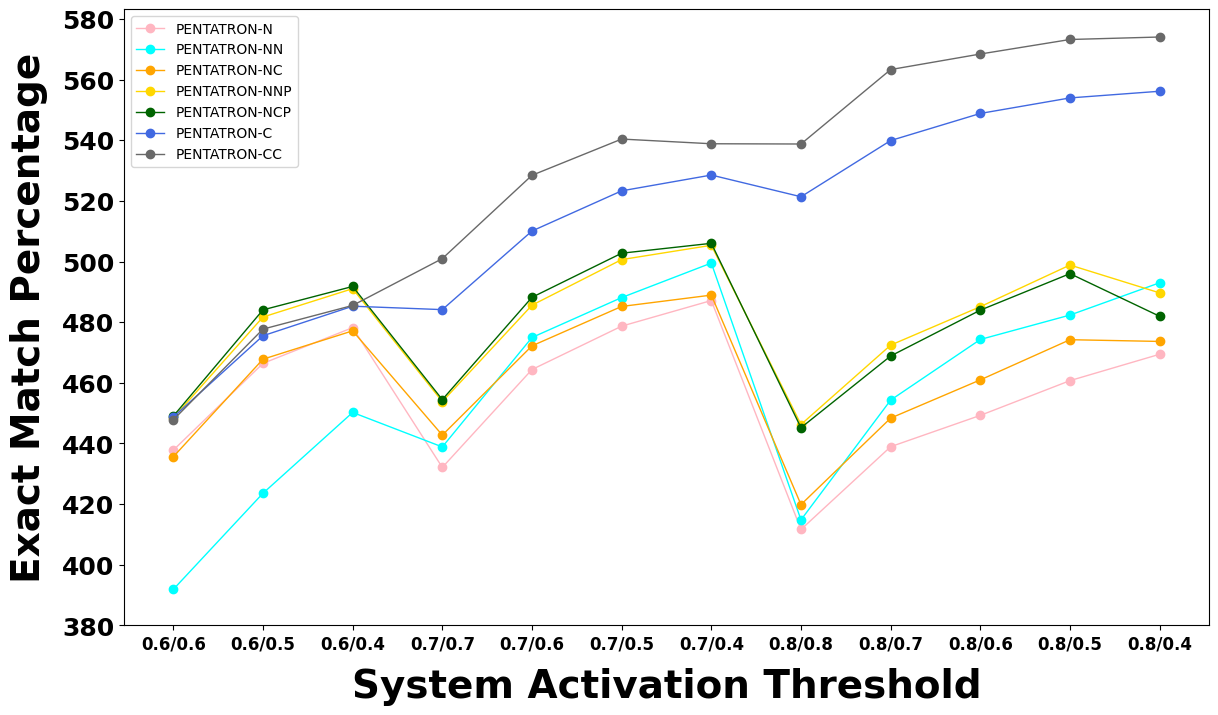}
    \caption{Performance of different versions of PENTATRON with respect to different system activation thresholds $\tau_1$ and $\tau_2$.}
    \label{fig:accuracy_vs_threshold}
\end{figure}

% \begin{figure}[h]
%     \centering
%     \includegraphics[width=\columnwidth]{emnlp_2022/figures/trigger_rate_vs_threshold.png}
%     \caption{We study the trigger rate of }
%     \label{fig:thresholding}
% \end{figure}

%\subsection{Results}
%----tables

%\subsection{Qualitative Analysis}

%\subsection{Domain Analysis}

\subsection{Observations and Case-studies}
%%%%%%%%%%%%%%%%%%%%%%%%%%%%%%%%%%%%%%%%%%%%%
% table: Case studies for contextual information
\makeatletter
\def\hlinewd#1{%
\noalign{\ifnum0=`}\fi\hrule \@height #1 \futurelet
\reserved@a\@xhline}
\makeatother

\newcommand*\colourcheck[1]{%
  \expandafter\newcommand\csname #1check\endcsname{\textcolor{#1}{\ding{52}}}%
}
\colourcheck{blue}
\colourcheck{green}
\colourcheck{red}

\newcommand*\colourcross[1]{%
  \expandafter\newcommand\csname #1cross\endcsname{\textcolor{#1}{\ding{55}}}%
}
\colourcross{blue}
\colourcross{green}
\colourcross{red}

\begin{table*}[!ht]
\centering

\scriptsize

\makebox[\textwidth][c]{
\begin{tabular}{l|l|l}
\hlinewd{1pt}
Dialogue Context         &  \begin{tabular}[l]{@{}l@{}} {[USER]} Turn on ben's light. \\ {[DEVICE]} I'm sorry I couldn't find the device. \\ {[USER]} Turn on benny's light. \\ {[DEVICE]} Okay. \\ \end{tabular} &  \begin{tabular}[l]{@{}l@{}} {[USER]} Play calen playlist. \\ {[DEVICE]} I could not find that on Amazon Music. \\ {[USER]} Play scars. \\ {[DEVICE]} Here’s Scars , by James Bay , on Amazon Music. \\ \end{tabular} \\ \hline
User Query           &  Turn \textcolor{orange}{ben's light} on pink & Play playlist \textcolor{orange}{karen} \\ \hline 
Rewrite Label            & Turn \textcolor{blue}{benny's light} on pink &  Play playlist \textcolor{blue}{callen} \\ \hlinewd{1pt}
DPR-EC & Turn \textbf{brecken's light} on pink \redcross & Play playlist \textbf{cameron} \redcross \\ \hline
PENTATRON-N             & Turn \textbf{britney's light} on pink  \redcross &  Play playlist \textbf{carrie} \redcross \\ \hline
PENTATRON-C & Turn \textbf{benny's light} on pink \greencheck & Play playlist \textbf{carrie} \redcross \\ \hline
PENTATRON-CC             & Turn \textbf{benny's light} on pink \greencheck & Play playlist \textbf{callen} \greencheck \\ \hlinewd{1pt}
\end{tabular}
}

\caption{Two examples to showcase the importance of full contextualization and personalization.}
\label{tab:contextual_case_study}
\end{table*}

Table~\ref{tab:ablations} shows the main results of our different versions of systems. The experimental result is consistent with our intuition. Since the pipeline has also actually been run on live traffic, through an A/B experiment (section~\ref{sec:ab}), the baselines were created for the purpose of this paper. All the \name{} models benefit from a personalization settings and outperform a global-wise retrieval model DPR-EC by a large margin. Among different settings of \name{}, it's obvious that both contextualization and multi-tasking bring further improvement. There is also some gain by adding task markers in the multi-task settings.

Figure~\ref{fig:thresholding} presents an ablation that shows the benefits of hard negative sampling. To further stress test our system, we also swept over different thresholds, summarized in Figure~\ref{fig:accuracy_vs_threshold}. We could notice that the general trend is consistent using different thresholds.

\begin{figure}[h]
    \centering
    \includegraphics[width=\columnwidth]{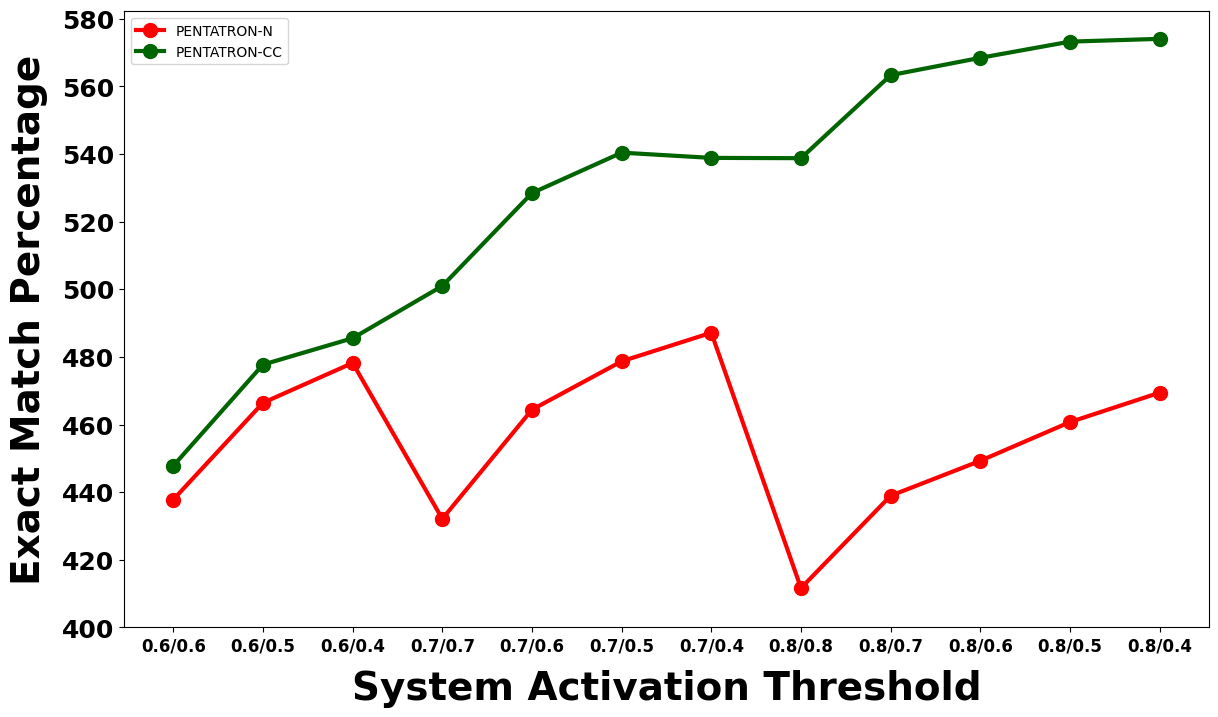}
    \caption{Demonstrating the value of contextual information with appropriate multitasking design.}
    \label{fig:accuracy_vs_treshold_best}
\end{figure}

In sweeping across thresholds in our empirical studies (Figure~\ref{fig:accuracy_vs_treshold_best}), we observe interesting trends. In particular, that when $\tau_1 = \tau_2$, the personalized model that does not utilize contextual information suffers from noisy predictions when the thresholds are equal since the top-$2$ retrieved entities are semantically very similar and the model finds it difficult to disambiguate. However, with the contextual information, we see consistent improvements in accuracy as we tighten thresholds.

%We analyze the logs of \name{} and present a few examples to further illustrate the benefits of full contextualization and personalization with respect to entity correction. This is described in Table~\ref{tab:contextual_case_study}.
We illustrate the benefits of our approach on a generic dialog in Table 3. In the left example from \textit{HomeAutomation} domain, the device name in the source query is incorrect which will make this task-oriented dialogue system fail. \name-C and \name-CC could generate the correct rewrite by leveraging dialogue context and user's personalized index which contains user's registered device name. A similar trend can be observed in the right example from Music domain. Besides, the right example also illustrates the benefits from multi-task learning by comparing the prediction from \name-C and \name-CC. Both the video name `carrie' and playlist name `callen' exist in user's personalized index. With the help of contrastive representation learning, \name-CC could learn to retrieve a Music domain entity which is the correct one here.

\begin{figure}[h]
    \centering
    \includegraphics[width=\columnwidth]{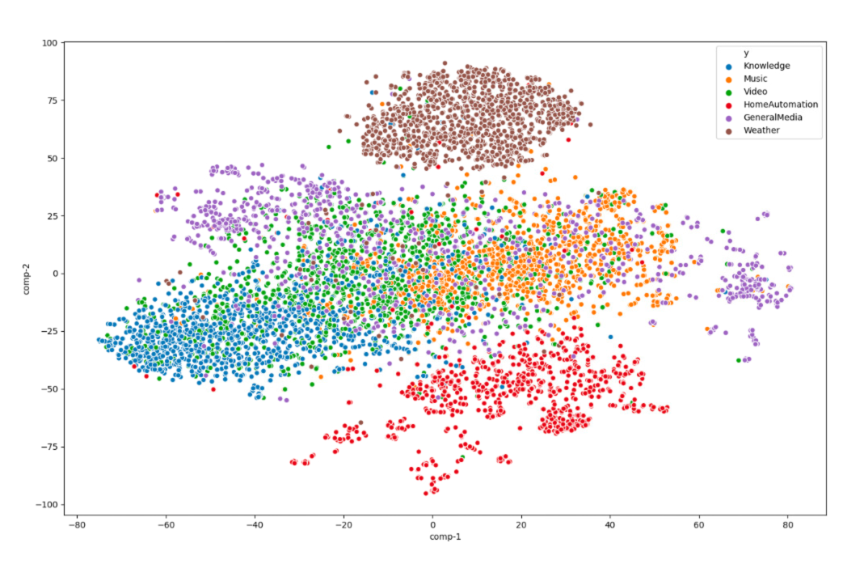}
    \caption{In the absence of the auxiliary task, queries across domains are interspersed which leads to lower accuracy due ambiguity in the rewrite domain. Here, the blue cluster denotes Knowledge domain queries, the orange cluster denotes Music domain queries and the green cluster denotes Video domain queries.}
    \label{fig:non_clustered}
\end{figure}

\begin{figure}[h]
    \centering
    \includegraphics[width=\columnwidth]{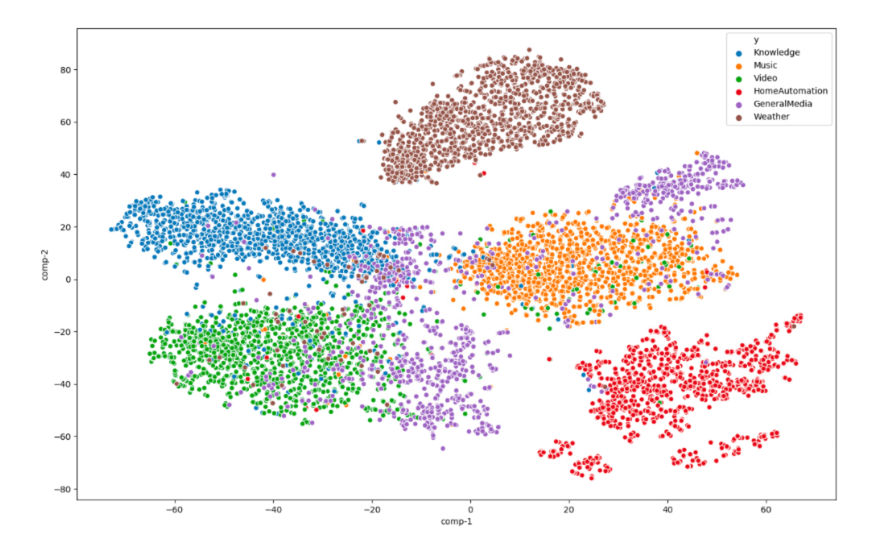}
    \caption{Multi-tasking to predict the rewrite domain, in addition to predicting the correct entity, leads to higher accuracy due to domain disambiguation arising from the implicit clustering effect.}
    \label{fig:clustered}
\end{figure}

\paragraph{Visualization: }We analyze the benefits of our design using t-SNE~\cite{van2008visualizing}. The results are presented in Figures \ref{fig:non_clustered} and \ref{fig:clustered}. We clearly observe that multi-tasking enables domain disambiguation via implicitly clustering the queries by domains, thus contributing positively to entity prediction accuracy and, in turn, improving the query rewrite quality. In particular, we observe that Music, Video and Knowledge domains immensely benefit from multi-tasking.


\subsection{Online Performance}
\label{sec:ab}
\paragraph{A/B Experimentation: }At the time of writing this, we deployed a static (request, rewrite) look-up table computed using \name-N to serve real users. With a $p$-value $< 0.05$, we observe a significant improvement, of 47.5\%, in the user experience measured using the model-based~\cite{gupta2021robertaiq} assessment used for dataset selection in Section~\ref{sec:train_test_data} on the treatment group as compared to the control group. Moreover, other friction metrics such as the turn error rate have improved over 40\% throughout the A/B duration. Successive version upgrade deployments are ongoing.

\paragraph{Latency: }To investigate the deployment in a real-time inference service, we performed extensive load tests implemented with a Flask endpoint. We store all objects in the main memory. On a c5.9x-large instance on AWS cloud, at $120$ queries per second hitting the \name{} system, we observed a P90 latency of less than $30$ms for the end-to-end execution.

\section{Conclusions and Future Directions}
In this work, we build a system called \name{} which significantly improves user experience in intelligent devices by operating on entities and reducing friction in multi-turn dialogues. There are several future directions we plan to work on, including operationalizing large-scale unbiased personalized and context-aware systems, and designing self-learning~\cite{ponnusamy2020feedback,roshan2020personalized} using techniques such as reinforcement learning. We also plan to investigate the utility of a multi-level index to improve entity coverage and mitigate the cold-start problem for new customers. Dynamic index building and deployment in low-latency applications is an ongoing direction.

%As future directions, we aim to augment the self-learning part.

% \newpage
\section*{Limitations}
\label{sec:limitations}
Our system has the following limitations. Though personalization offers great benefits, the coverage of desired entities in our historical index due to personalization is typically limited. Specifically, we observe only 20\% coverage in our empirical studies. This can alleviated using a multi-level index involving clusters of users. We have initial results on this approach and plan to compile that in future work.

Next, natural language based prompts should further improve our system. However, very long sequence length has concerns with respect latency and memory on CPU-deployed solutions. A potential solution to this is to consider low-rank factorization in the attention design.

Finally, in production deployments, large-scale in-memory index for multiple locales poses cost challenges. A separate study is warranted to study hybrid storage mechanisms and high performance cache design.

%EMNLP 2022 requires all submissions to have a section titled ``Limitations'', for discussing the limitations of the paper as a complement to the discussion of strengths in the main text. This section should occur after the conclusion, but before the references. It will not count towards the page limit.  

%The discussion of limitations is mandatory. Papers without a limitation section will be desk-rejected without review. While we are open to different types of limitations, just mentioning that a set of results have been shown for English only probably does not reflect what we expect. Mentioning that the method works mostly for languages with limited morphology, like English, is a much better alternative. In addition, limitations such as low scalability to long text, the requirement of large GPU resources, or other things that inspire crucial further investigation are welcome.

\section*{Ethics Statement}
\label{sec:ethics}
To the best of our knowledge, our work is ethical and has a positive impact on society and human well-being. In particular, we take pride in emphasizing that we handle customer confidentiality and privacy with critical care. Its design principles are unbiased.
% Scientific work published at EMNLP 2022 must comply with the \href{https://www.aclweb.org/portal/content/acl-code-ethics}{ACL Ethics Policy}. We encourage all authors to include an explicit ethics statement on the broader impact of the work, or other ethical considerations after the conclusion but before the references. The ethics statement will not count toward the page limit (8 pages for long, 4 pages for short papers).

% \section*{Acknowledgements}

\bibliography{custom}
\bibliographystyle{acl_natbib}

\end{document}